\documentclass[11pt,a4paper]{article}
\usepackage[hyperref]{acl2021}
\usepackage{times}
\usepackage{latexsym}

\usepackage{microtype}
\usepackage{multirow}
\usepackage{amsfonts}
\usepackage{graphicx}
\usepackage{amsmath}
\usepackage{subfigure}
\aclfinalcopy 
\def\aclpaperid{596} 

\title{Towards Propagation Uncertainty: Edge-enhanced Bayesian Graph Convolutional Networks for Rumor Detection}

\author{Lingwei Wei$^{1,4}$, Dou Hu$^{2}$, Wei Zhou$^{1*}$, Zhaojuan Yue$^{3}$, Songlin Hu$^{1,4*}$ \\
$^1$ Institute of Information Engineering, Chinese Academy of Sciences \\
$^2$ National Computer System Engineering Research Institute of China \\ 
$^3$ Computer Network Information Center, Chinese Academy of Sciences \\ 
$^4$ School of Cyber Security, University of Chinese Academy of Sciences \\
  \texttt{\{weilingwei18, hudou18\}@mails.ucas.edu.cn}  \\ 
  \texttt{\{zhouwei, husonglin\}@iie.ac.cn} \\
  \texttt{yuezhaojuan@cnic.cn} \\
  }

\date{}

\begin{document}
\maketitle

\begin{abstract}
\let\thefootnote\relax\footnotetext{* Corresponding author.}

Detecting rumors on social media is a very critical task with significant implications to the economy, public health, etc. Previous works generally capture effective features from texts and the propagation structure. 
However, the {\it uncertainty} caused by unreliable relations in the propagation structure is common and inevitable due to wily rumor producers and the limited collection of spread data. Most approaches neglect it and may seriously limit the learning of features. 
Towards this issue, this paper makes the first attempt to explore propagation uncertainty for rumor detection. Specifically, we propose a novel {\bf E}dge-enhanced {\bf B}ayesian {\bf G}raph {\bf C}onvolutional {\bf N}etwork ({\bf EBGCN}) to capture robust structural features. The model adaptively rethinks the reliability of latent relations by adopting a Bayesian approach. Besides, we design a new edge-wise consistency training framework to optimize the model by enforcing consistency on relations. Experiments on three public benchmark datasets demonstrate that the proposed model achieves better performance than baseline methods on both rumor detection and early rumor detection tasks.
\end{abstract}

\section{Introduction}
With the ever-increasing popularity of social media sites, user-generated messages can quickly reach a wide audience.
However, social media can also enable the spread of false rumor information \cite{Vosoughi1146}.
Rumors are now viewed as one of the greatest threats to democracy, journalism, and freedom of expression. 
Therefore, detecting rumors on social media is highly desirable and socially beneficial \cite{DBLP:journals/osnm/AhsanKS19}.

Almost all the previous studies on rumor detection leverage text content including the source tweet and all user retweets or replies.
As time goes on, rumors form their specific propagation structures after being retweeted or replied to. 
\citet{article2015,Vosoughi1146} have confirmed rumors spread significantly farther, faster, deeper, and more broadly than the truth.
They provide the possibility of detecting rumors through the propagation structure.
Some works \cite{DBLP:conf/ijcai/MaGMKJWC16,DBLP:conf/coling/KochkinaLZ18} typically learn temporal features alone from propagation sequences, ignoring the internal topology.
Recent approaches \cite{DBLP:conf/acl/WongGM18,DBLP:conf/aaai/KhooCQ020} model the propagation structure as trees to capture structural features. 
\citet{DBLP:conf/aaai/BianXXZHRH20,DBLP:conf/emnlp/WeiXM19} construct graphs and aggregate neighbors' features through edges based on reply or retweet relations.

However, most of them only work well in a narrow scope since they treat these relations as reliable edges for message-passing. 
\begin{figure}[t]
  \centering
  \includegraphics[width=0.99\linewidth]{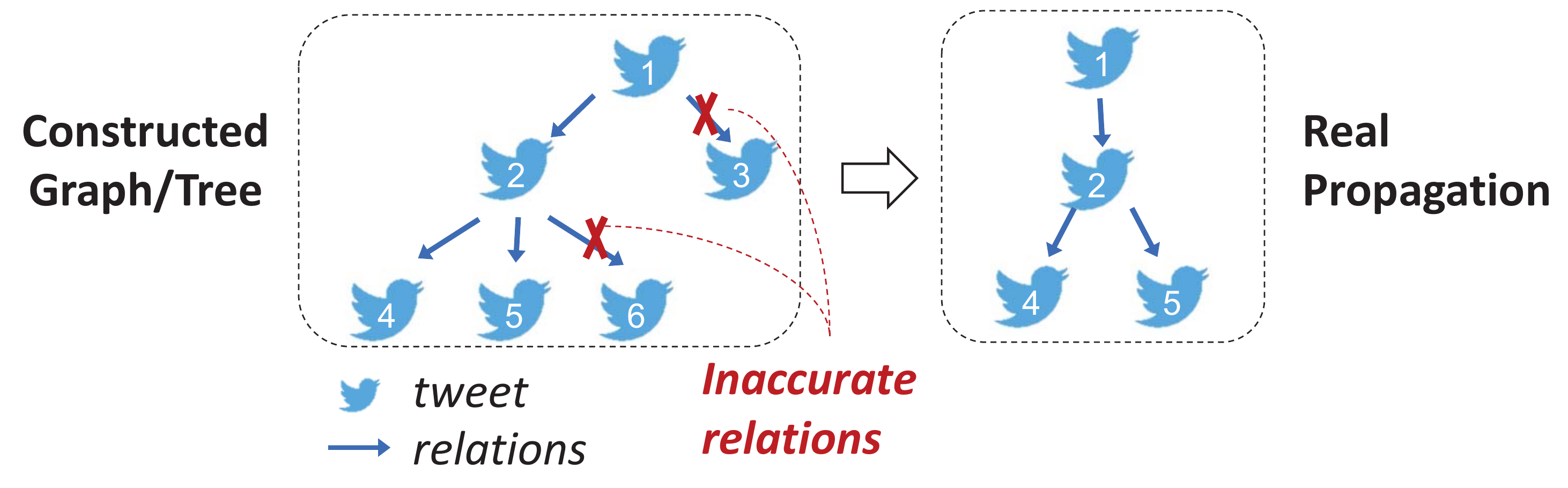}
  \caption{An example of uncertain propagation structure. It includes inaccurate relations, making constructed graph inconsistent with the real propagation.  }
  \label{fig:example}
\end{figure}
As shown in Figure \ref{fig:example}, the existence of inaccurate relations brings {\it uncertainty} in the propagation structure. 
The neglect of unreliable relations would lead to severe error accumulation through multi-layer message-passing and limit the learning of effective features.

We argue such inherent uncertainty in the propagation structure is inevitable for two aspects: i) In the real world, rumor producers are always wily. They tend to viciously manipulate others to create fake supporting tweets or remove opposing voices to evade detection \cite{DBLP:conf/ijcai/YangLTLLZ20}. 
In these common scenarios, relations can be manipulated, which provides uncertainty in the propagation structure.
ii) Some annotations of spread relations are subjective and fragmentary \cite{DBLP:conf/acl/MaGW17,DBLP:journals/corr/ZubiagaHLPT15}. 
The available graph would be a portion of the real propagation structure as well as contain noisy relations, resulting in uncertainty.
Therefore, it is very challenging to handle inherent uncertainty in the propagation structure to obtain robust detection results.

To alleviate this issue, we make the first attempt to explore the uncertainty in the propagation structure.
Specifically, we propose a novel {\bf E}dge-enhanced {\bf B}ayesian {\bf G}raph {\bf C}onvolutional {\bf N}etwork ({\bf EBGCN}) for rumor detection to model the uncertainty issue in the propagation structure from a probability perspective.
The core idea of EBGCN is to adaptively control the message-passing based on the prior belief of the observed graph to surrogate the fixed edge weights in the propagation graph.
In each iteration, edge weights are inferred by the posterior distribution of latent relations according to the prior belief of node features in the observed graph.
Then, we utilize graph convolutional layers to aggregate node features by aggregating various adjacent information on the refining edges.
Through the above network, EBGCN can handle the uncertainty in the propagation structure and promote the robustness of rumor detection.

Moreover, due to the unavailable of missing or inaccurate relations for training the proposed model, we design a new edge-wise consistency training framework.
The framework combines unsupervised consistency training on these unlabeled relations into the original supervised training on labeled samples, to promote better learning. 
We further ensure the consistency between the latent distribution of edges and the distribution of node features in the observed graph by computing KL-divergence between two distributions.
Ultimately, both the cross-entropy loss of each claim and the Bayes by Backprop loss of latent relations will be optimized to train the proposed model.

We conduct experiments on three real-world benchmark datasets ({\it i.e.,} {\it Twitter15}, {\it Twitter16}, and {\it PHEME}).  
Extensive experimental results demonstrate the effectiveness of our model. 
EBGCN offers a superior uncertainty representation strategy and boosts the performance for rumor detection. The main contributions of this work are summarized as follows:
\begin{itemize}
    \item We propose novel Edge-enhanced Bayesian Graph Convolutional Networks (EBGCN) to handle the uncertainty in a probability manner.
    To the best of our knowledge, this is the first attempt to consider the inherent uncertainty in the propagation structure for rumor detection.
    \item We design a new edge-wise consistency training framework to optimize the model with unlabeled latent relations.
    \item Experiments on three real-world benchmark datasets demonstrate the effectiveness of our model on both rumor detection and early rumor detection tasks\footnote{The source code is available at \url{https://github.com/weilingwei96/EBGCN}.}.  
\end{itemize}

\section{Related Work}
\subsection{Rumor Detection}
Traditional methods on rumor detection adopted machine learning classifiers based on handcrafted features, such as sentiments \cite{DBLP:conf/www/CastilloMP11}, bag of words \cite{DBLP:conf/semeval/EnayetE17} and time patterns \cite{DBLP:conf/cikm/MaGWLW15}. 
Based on salient features of rumors spreading,
\citet{DBLP:conf/icde/WuYZ15,DBLP:conf/acl/MaGW17} modeled propagation trees and then used SVM with different kernels to detect rumors. 

Recent works have been devoted to deep learning methods. 
\citet{DBLP:conf/ijcai/MaGMKJWC16} employed Recurrent Neural Networks (RNN) to sequentially process each timestep in the rumor propagation sequence.
To improve it, many researchers captured more long-range dependency via attention mechanisms \cite{DBLP:conf/pakdd/ChenLYZ18}, convolutional neural networks \cite{DBLP:conf/ijcai/YuLWWT17,DBLP:conf/cikm/ChenSHG19}, and Transformer \cite{DBLP:conf/aaai/KhooCQ020}.
However, most of them focused on learning temporal features alone, ignoring the internal topology structure.

To capture topological-structural features, \citet{DBLP:conf/acl/WongGM18} presented two recursive neural network (RvNN) based on bottom-up and top-down propagation trees.
\citet{DBLP:conf/icdm/YuanMZHH19,DBLP:conf/acl/LuL20,DBLP:conf/cikm/NguyenSNK20} formulated the propagation structure as graphs.
Inspired by Graph Convolutional Network (GCN) \cite{DBLP:conf/iclr/KipfW17}, 
\citet{DBLP:conf/aaai/BianXXZHRH20} first applied two GCNs based on the propagation and dispersion graphs.  
\citet{DBLP:conf/emnlp/WeiXM19} jointly modeled the structural property by GCN and the temporal evolution by RNN.

\begin{figure*}[t]
  \centering
  \includegraphics[width=0.99\linewidth]{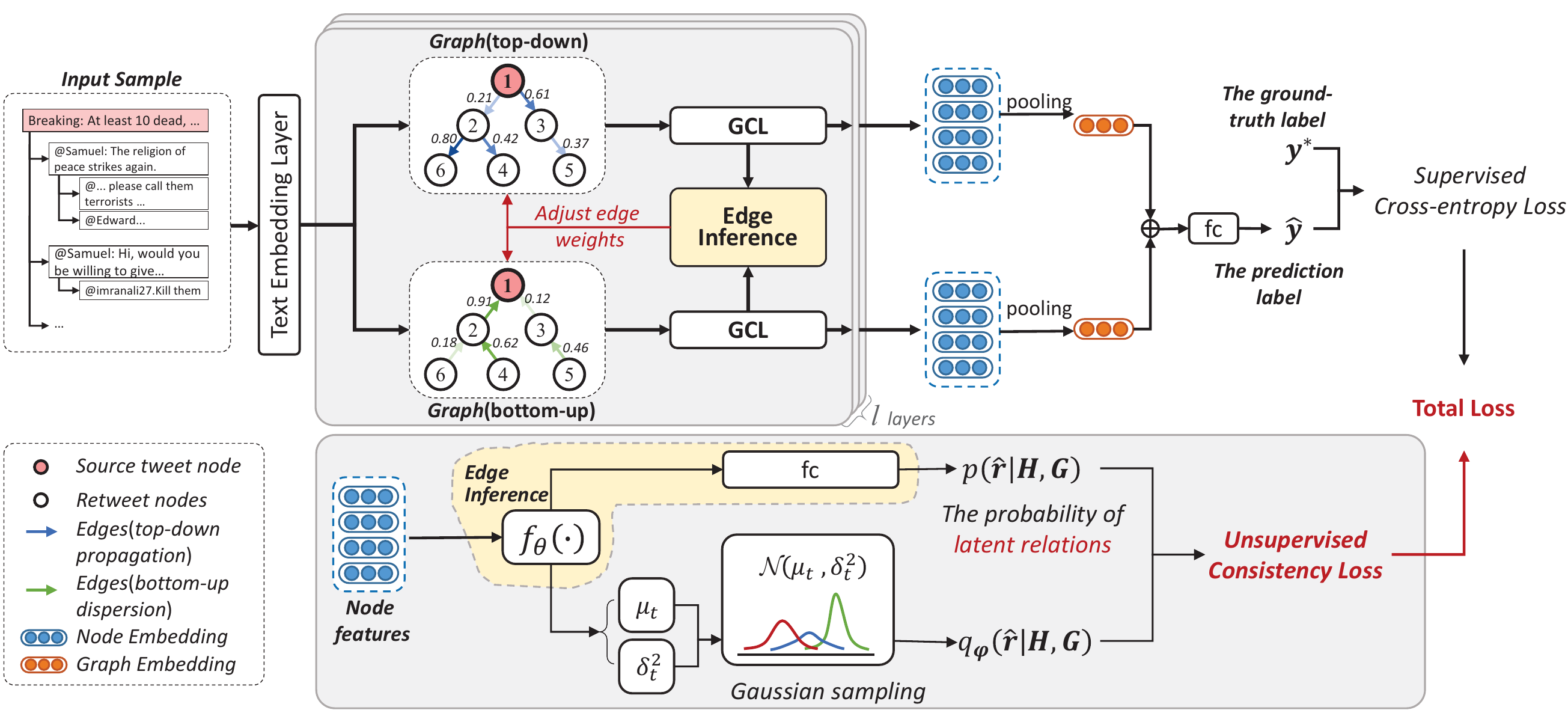}  
  \caption{The architecture of the proposed rumor detection model EBGCN.}
  \label{fig2:overview}
\end{figure*}

However, most of them treat the edge as the reliable topology connection for message-passing. 
Ignoring the uncertainty caused by unreliable relations could lead to lacking robustness and make it risky for rumor detection.
Inspired by valuable research \cite{DBLP:conf/www/ZhangLLY19} that modeled uncertainty caused by finite available textual contents, this paper makes the first attempt to consider the uncertainty caused by unreliable relations in the propagation structure for rumor detection.

\subsection{Graph Neural Networks}
Graph Neural Networks (GNNs) \cite{DBLP:conf/iclr/KipfW17, DBLP:conf/esws/SchlichtkrullKB18, DBLP:conf/iclr/VelickovicCCRLB18} have demonstrated remarkable performance in modeling structured data in a wide variety of fields, {\it e.g.}, text classifcation \cite{DBLP:conf/aaai/YaoM019}, recommendation system \cite{DBLP:conf/aaai/WuT0WXT19} and emotion recognition \cite{DBLP:conf/emnlp/GhosalMPCG19}. 
Although promising, they have limited capability to handle uncertainty in the graph structure.
While the graphs employed in real-world applications are themselves derived from noisy data or modeling assumptions. 
To alleviate this issue, some valuable works \cite{DBLP:conf/aaai/LuoHZWBY20,DBLP:conf/aaai/ZhangPCU19} provide an approach for incorporating uncertain graph information by exploiting a Bayesian framework \cite{DBLP:conf/nips/MaddoxIGVW19}.
Inspired by them, this paper explores the uncertainty in the propagation structure from a probability perspective, to obtain more robust rumor detection results.

\section{Problem Statement}
This paper develops EBGCN which processes text contents and propagation structure of each claim for rumor detection.
In general, \textbf{rumor detection} commonly can be regarded as a multi-classification task, which aims to learn a classifier from training claims for predicting the label of a test claim. 

Formally, let $\mathcal{C} = \{ c^1, c^2, ..., c^m \}$ be the rumor detection dataset, where $c^i$ is the $i$-th claim and $m$ is the number of claims. For each claim $c^i=\{ r^i, x^i_1, x^i_2, ..., x^i_{n_i-1}, G^i \}$, $G^i$ indicates the propagation structure, $r^i$ is the source tweet, $x^i_j$ refers to the $j$-th relevant retweet, and $n_i$ represents the number of tweets in the claim $c^i$.
Specifically, $G^i$ is defined as a propagation graph $ G_i = \langle V_i, E_i \rangle$ with the root node $r^i$ \cite{DBLP:conf/acl/WongGM18,DBLP:conf/aaai/BianXXZHRH20}, where $V_i = \{ r^i, x^i_1, x^i_2, ..., x^i_{n_i-1} \}$ refers to the node set and $E_i = \{ e^i_{st} | s,t=0,...,n_i-1  \}$ represent a set of directed edges from a tweet to its corresponding retweets. 
Denote $\textbf{A}_i \in \mathbb{R}^{n_i \times n_i}$ as an adjacency matrix where the initial value is 
$$
\alpha_{st}  = \left\{\begin{matrix}
  1, & \text{if} \  e^i_{st} \in E_i  \\ 
  0, & \text{otherwise}
  \end{matrix} \right..
$$ 

Besides, each claim $c^i$ is annotated with a ground-truth label $y^i \in \mathcal{Y}$, where $\mathcal{Y}$ represents fine-grained classes.
Our goal is to learn a classifier from the labeled claimed set, that is $f: \mathcal{C} \rightarrow \mathcal{Y}$.

\section{The Proposed Model}
In this section, we propose a novel edge-enhanced bayesian graph convolutional network (EBGCN) for rumor detection in Section~\ref{sec:ebgcn}.
For better training, we design an edge-wise consistency training framework to optimize EBGCN in Section~\ref{sec:train}.

\subsection{Overview}
The overall architecture of EBGCN is shown in Figure \ref{fig2:overview}.
Given the input sample including text contents and its propagation structure, we first formulate the propagation structure as directed graphs with two opposite directions, {\it i.e.,} a top-down propagation graph and a bottom-up dispersion graph.
Text contents are embedded by the text embedding layer. 
After that, we iteratively capture rich structural characteristics via two main components, node update module, and edge inference module.
Then, we aggregate node embeddings to generate graph embedding and output the label of the claim.

For training, we incorporate unsupervised consistency training on the Bayes by Backprop loss of unlabeled latent relations.
Accordingly, we optimize the model by minimizing the weighted sum of the unsupervised loss and supervised loss.

\subsection{Edge-enhanced Bayesian Graph Convolutional Networks} \label{sec:ebgcn}
\subsubsection{Graph Construction and Text Embedding}

The initial graph construction is similar to the previou work \cite{DBLP:conf/aaai/BianXXZHRH20}, {\it i.e.,} build two distinct directed graphs for the propagation structure of each claim $c^i$. The top-down propagation graph and bottom-up dispersion graph are denoted as $G^{TD}_i$ and $G^{BU}_i$, respectively. Their corresponding initial  adjacency matrices are $\textbf{A}^{TD}_i=\textbf{A}_i$ and $\textbf{A}^{BU}_i=\textbf{A}_i^\top$.
Here, we leave out the superscript $i$ in the following description for better presenting our method.

The initial feature matrix of postings in the claim $c$ can be extracted Top-5000 words in terms of \textit{TF-IDF} values, denoted as $\textbf{X}=[ \textbf{x}_0, \textbf{x}_1, ..., \textbf{x}_{n-1} ] \in \mathbb{R}^{n \times d_0}$,
where $\textbf{x}_0 \in \mathbb{R}^{d_0} $ is the vector of the source tweet and $d_0$ is the dimensionality of textual features. 
The initial feature matrices of nodes in propagation graph and dispersion graph are the same, {\it i.e.,} $\textbf{X}^{TD}=\textbf{X}^{BU}=\textbf{X}$.

\subsubsection{Node Update}
Graph convolutional networks (GCNs) \cite{DBLP:conf/iclr/KipfW17} are able to extract graph structure information and better characterize a node's neighborhood.
They define multiple Graph Conventional Layers (GCLs) to iteratively aggregate features of neighbors for each node and can be formulated as a simple differentiable message-passing framework.
Motivated by GCNs, we employ the GCL to update node features in each graph.
Formally, node features at the $l$-th layer $\textbf{H}^{(l)} = [\textbf{h}^{(l)}_0, \textbf{h}^{(l)}_1, ..., \textbf{h}^{(l)}_{n-1} ]$ can be defined as,
\begin{equation}
  \textbf{H}^{(l)} = \sigma (\hat{\textbf{A}}^{(l-1)} \textbf{H}^{(l-1)} \textbf{W}^{(l)} + \textbf{b}^{(l)}),
\end{equation}
where $\hat{\textbf{A}}^{(l-1)}$ represents the normalization of adjacency matrix $\textbf{A}^{(l-1)}$ \cite{DBLP:conf/iclr/KipfW17}.
We initialize node representations by textual features, {\it i.e.,} $\textbf{H}^{(0)}=\textbf{X}$.

\subsubsection{Edge Inference}
To alleviate the negative effects of unreliable relations, we rethink edge weights based on the currently observed graph by adopting a soft connection.

Specifically, we adjust the weight between two nodes by computing a transformation ${f}_{e}(\cdot; \theta_t)$ based on node representations at the previous layer. 
Then, the adjacency matrix will be updated, {\it i.e.,} 
\begin{equation}
  \begin{split}
    \textbf{g}_t^{(l)} &=  f_e\left( \| \textbf{h}^{(l-1)}_i -  \textbf{h}^{(l-1)}_j \| ; {\theta}_t \right) , \\
    \textbf{{A}}^{(l)}  &= \sum\limits_{t=1}^T \sigma(\textbf{W}_t^{(l)} \textbf{g}_t^{(l)} + \textbf{b}_t^{(l)}) \cdot \textbf{{A}}^{(l-1)}.\label{eq:T}
\end{split}
\end{equation}
In practice, ${f}_{e}(\cdot; \theta_t)$ consists an convolutional layer and an activation function. 
$T$ refers to the number of latent relation types.
$\sigma(\cdot)$ refers to a {\it sigmoid} function. 
$\textbf{W}^{(l)}_t$ and $\textbf{W}^{(l)}_t$ are learnable parameters.

We perform share parameters to the edge inference layer in two graphs $G^{TD}$ and $G^{BU}$. 
After the stack of transformations in two layers, the model can effectively accumulate a normalized sum of features of the neighbors driven by latent relations, denoted as ${\textbf{H}}^{TD}$ and  ${\textbf{H}}^{BU}$.

\subsubsection{Classification}

We regard the rumor detection task as a graph classification problem.
To aggregate node representations in the graph, we employ aggregator to form the graph representations. Given the node representations in the propagation graph ${\textbf{H}}^{TD}$ and the node representations in the dispersion graph ${\textbf{H}}^{BU}$, the graph representations can be computed as:
\begin{equation}
  \begin{split}
    \textbf{C}^{TD} &= {meanpooling}({\textbf{H}}^{TD}), \\
    \textbf{C}^{BU} &= {meanpooling}({\textbf{H}}^{BU}),
  \end{split}
\end{equation}
where ${meanpooling}(\cdot)$ refers to the mean-pooling aggregating function. 
Based on the concatenation of two distinct graph representations, label probabilities of all classes can be defined by a full connection layer and a softmax function,  {\it i.e.,} 
\begin{equation}
    \hat {\textbf{y}} = softmax\left(\textbf{W}_c [ \textbf{C}^{TD} ; \textbf{C}^{BU} ] + \textbf{b}_c\right),
\end{equation}
where $\textbf{W}_c$ and $\textbf{b}_c$ are learnable parameter matrices.

\subsection{Edge-wise Consistency Training Framework} \label{sec:train}

{\bf For the supervised learning loss} $\mathcal{L}_{c}$, we compute the cross-entropy of the predictions and ground truth distributions $C=\{c_1, c_2, ..., c_m\}$, {\it i.e.,} 
\begin{equation}
    \mathcal{L}_c = - \sum_{i}^{|\mathcal{Y}|}   \textbf{y}^i log {\hat{\textbf{y}}^i}, 
\end{equation}
where $\textbf{y}^i $ is a vector representing distribution of ground truth label for the $i$-th claim sample.

{\bf For the unsupervised learning loss} $\mathcal{L}_e$, we amortize the posterior distribution of the classification weight $p(\varphi)$ as $q(\varphi)$ to enable quick prediction at the test stage and learn parameters by minimizing the average expected loss over latent relations, {\it i.e.,} $\varphi^* = \text{arg} \min_{\varphi} \mathcal{L}_e$, where 
\begin{equation}
    \resizebox{0.89\linewidth}{!}{$
  \begin{split}
  \mathcal{L}_{e} &= \mathbb{E} \left[ D_{KL}\left(
      p ( 
          \hat{\textbf{r}}^{(l)}  |\textbf{H}^{(l-1)},G
          ) 
      \|  
      q_{\varphi} (
          \hat{\textbf{r}}^{(l)}|\textbf{H}^{(l-1)},G
          )
      \right)
      \right],
      \\
      \varphi^* &= \text{arg} \max\limits_{\varphi} \mathbb{E}
      [
       \log \int p(\hat{\textbf{r}}^{(l)} | \textbf{H}^{(l-1)}, \varphi ) 
       q_{\varphi} (
           \varphi | \textbf{H}^{(l-1)},G 
           ) d\varphi 
           ],
  \end{split}
    $}
\end{equation}
where $\hat{\textbf{r}}$ is the prediction distribution of latent relations.
To ensure likelihood tractably, we model the prior distribution of each latent relation $r_t, t\in[1,T]$ independently. For each relation, we define a factorized Gaussian distribution for each latent relation $q_{\varphi} (
    \varphi |\textbf{H}^{(l-1)},G; \Theta)$ with means $\mu_t$ and variances $\delta^2_t$ set by the transformation layer,
\begin{equation}
  \resizebox{0.89\linewidth}{!}{$
  \begin{split}
    q_{\varphi} (
        \varphi |\textbf{H}^{(l-1)},G
        ; \Theta)) &=  \prod \limits_{t=1}^{T}  q_{\varphi} (\varphi_t | \{{\textbf{g}_{t}^{(l)} }  \} _{t=1}^{T} )  \\ &= \prod \limits_{t=1}^{T} \mathcal{N}(\mu_t, \delta_t^2), \\
    \mu_t =  f_{\mu}(\{  \textbf{g}_{t}^{(l)}   \}_{t=1}^{T}&;\theta_{\mu}),  
    \delta^2_t = f_{\delta}(\{{ \textbf{g}_{t}^{(l)} \}}_{t=1}^{T};\theta_{\delta}),
  \end{split}
  $}
\end{equation}
where $f_{\mu}(\cdot;\theta_{\mu})$ and $f_{\delta}(\cdot;\theta_{\mu})$ refer to compute the mean and variance of input vectors, parameterized by $\theta_{\mu}$ and $\theta_{\delta}$, respectively. 
Such that amounts to set the weight of each latent relation.

Besides, we also consider the likelihood of latent relations when parameterizing the posterior distribution of prototype vectors. The likelihood of latent relations from the $l$-th layer based on node embeddings can be adaptively computed by,
\begin{equation}  
  \resizebox{0.89\linewidth}{!}{$
    \begin{split}
    p( \hat{\textbf{r}}^{(l)} | \textbf{H}^{(l-1)}, \varphi ) 
    &= \prod \limits_{t=1}^{T} p( 
      \hat{\textbf{r}}^{(l)}_{t} 
      |
      \textbf{H}^{(l-1)}, \varphi_t), \\
     p(\hat{\textbf{r}}^{(l)}_t|
      \textbf{H}^{(l-1)}, \varphi_t) &= 
      \frac{exp \left( 
        \textbf{W}_t  \textbf{g}^{(l)}_t + \textbf{b}_t 
        \right)} 
        {\sum_{t=1}^{T} exp\left(
            \textbf{W}_t  \textbf{g}^{(l)}_t + \textbf{b}_t
            \right)
            }.
\end{split}
$}
\end{equation}
In this way, the weight of edges can be adaptively adjusted based on the observed graph, which can thus be used to effectively pass messages and learn more discriminative features for rumor detection.

To sum up, in training, we optimize our model EBGCN by minimizing the cross-entropy loss of labeled claims $\mathcal{L}_{c}$ and Bayes by Backprop loss of unlabeled latent relations $\mathcal{L}_{e}$, {\it i.e.,}
\begin{equation}
    \Theta^* = \text{arg}\min \limits_{\Theta}   \gamma \mathcal{L}_{c} +  (1-\gamma) \mathcal{L}_{e}, 
  \end{equation}
  where $\gamma$ is the trade-off coefficient.

\section{Experimental Setup}

\subsection{Datasets}
We evaluate the model on three real-world benchmark datasets: {\it Twitter15} \cite{DBLP:conf/acl/MaGW17}, {\it Twitter16} \cite{DBLP:conf/acl/MaGW17}, and {\it PHEME} \cite{DBLP:journals/corr/ZubiagaHLPT15}.  The statistics are shown in Table \ref{tab:datasets}.
    \textbf{Twitter15} and \textbf{ Twitter16}\footnote{\url{https://www.dropbox.com/s/7ewzdrbelpmrnxu/rumdetect
    2017.zip?dl=0}} contain 1,490 and 818 claims, respectively. Each claim is labeled as Non-rumor (NR), False Rumor (F), True Rumor (T), or Unverified Rumor (U). 
    Following \cite{DBLP:conf/acl/WongGM18,DBLP:conf/aaai/BianXXZHRH20}, we randomly split the dataset into five parts and conduct 5-fold cross-validation to obtain robust results. 
    \textbf{PHEME} dataset\footnote{\url{https://figshare.com/articles/dataset/PHEME_dataset_for_Rumour_Detection_and_Veracity_Classification/6392078}} provides 2,402 claims covering nine events and contains three labels, False Rumor (F), True Rumor (T), and Unverified Rumor (U).
    Following the previous work \cite{DBLP:conf/emnlp/WeiXM19}, we conduct {leave-one-event-out} cross-validation, {\it i.e.,} in each fold, one event's samples are used for testing, and all the rest are used for training.
    

\begin{table}[t]
  \centering
  \resizebox{\linewidth}{!}{$
  \begin{tabular}{cccc}
      \hline
      Dataset & {\it Twitter15} & {\it Twitter16} & {\it PHEME} \\
      \hline
      \# of claims & 1,490  & 818 & 2,402 \\ 
      \# of false rumors & 370 & 205 & 638 \\ 
      \# of true rumors & 374 & 205 & 1,067 \\ 
      \# of unverified rumors & 374 & 203 & 697 \\ 
      \# of non-rumors & 372 & 205 & - \\
      \# of postings & 331,612 & 204,820 & 105,354 \\ 
      \hline
  \end{tabular}
  $}
  \caption{Statistics of the datasets.}
  \label{tab:datasets}
\end{table}

\subsection{Baselines}
For {\it Twitter15} and {\it Twitter16}, we compare our proposed model with the following methods.
     \textbf{DTC} \cite{DBLP:conf/www/CastilloMP11} adopted a decision tree classifier based on information credibility. 
 \textbf{SVM-TS} \cite{DBLP:conf/cikm/MaGWLW15} leveraged time series to model the chronological variation of social context features via a linear SVM classifier.
 \textbf{SVM-TK} \cite{DBLP:conf/acl/MaGW17} applied an SVM classifier with a propagation tree kernel to model the propagation structure of rumors. 
\textbf{GRU-RNN} \cite{DBLP:conf/ijcai/MaGMKJWC16} employed RNNs to model the sequential structural features.
 \textbf{RvNN} \cite{DBLP:conf/acl/WongGM18} adopted two recursive neural models based on a bottom-up and a top-down propagation tree. 
 \textbf{StA-PLAN} \cite{DBLP:conf/aaai/KhooCQ020} employed transformer networks to incorporate long-distance interactions among tweets with propagation tree structure.
 \textbf{BiGCN} \cite{DBLP:conf/aaai/BianXXZHRH20} utilized bi-directional GCNs to model bottom-up propagation and top-down dispersion.


For {\it PHEME}, we compare with several representative state-of-the-art baselines.
     \textbf{NileTMRG} \cite{DBLP:conf/semeval/EnayetE17} used linear support vector classification based on bag of words.
     \textbf{BranchLSTM} \cite{DBLP:conf/coling/KochkinaLZ18} decomposed the propagation tree into multiple branches and adopted a shared LSTM to capture structural features.
     \textbf{RvNN} \cite{DBLP:conf/acl/WongGM18} consisted of two recursive neural networks to model propagation trees.
     \textbf{Hierarchical GCN-RNN} \cite{DBLP:conf/emnlp/WeiXM19} modeled structural property based on GCN and RNN.
    \textbf{BiGCN} \cite{DBLP:conf/aaai/BianXXZHRH20} consisted of propagation and dispersion GCNs to learn structural features from propagation graph.

\subsection{Evaluation Metrics}
For {\it Twitter15} and {\it Twitter16}, we follow \cite{DBLP:conf/acl/WongGM18,DBLP:conf/aaai/BianXXZHRH20,DBLP:conf/aaai/KhooCQ020} and evaluate the \textbf{accuracy} (Acc.) over four categories and \textbf{F1} score ($F_1$) on each class. 
For {\it PHEME}, following \cite{DBLP:conf/semeval/EnayetE17,DBLP:conf/coling/KochkinaLZ18,DBLP:conf/emnlp/WeiXM19}, we apply the \textbf{accuracy} (Acc.),  \textbf{macro-averaged F1} (m$F_1$) as evaluation metrics. Also, we report the \textbf{weighted-averaged F1} (w$F_1$) because of the imbalanced class problem.

\subsection{Parameter Settings}
Following comparison baselines, the dimension of hidden vectors in the GCL is set to 64. 
The number of latent relations $T$ and the coefficient weight $\gamma$ are set to $[1,5]$ and $[0.0, 1.0]$, respectively.
we train the model via backpropagation and a wildly used stochastic gradient descent named Adam \cite{DBLP:journals/corr/KingmaB14}.
The learning rate is set to {$\{0.0002, 0.0005, 0.02\}$} for {\it Twitter15}, {\it Twitter16}, and {\it PHEME}, respectively.
The training process is iterated upon 200 epochs and early stopping \cite{2580217320070801early} is applied when the validation loss stops decreasing by $10$ epochs. 
The optimal set of hyperparameters are determined by testing the performance on the fold-$0$ set of {\it Twitter15} and {\it Twitter16}, and the class-balanced {\it charlie hebdo} event set of {\it PHEME}.

Besides, on {\it PHEME}, following \cite{DBLP:conf/emnlp/WeiXM19}, we replace TF-IDF features with word embeddings by skip-gram with negative sampling \cite{DBLP:conf/nips/MikolovSCCD13} and set the dimension of textual features to $200$. We implement this variant of BiGCN and EBGCN, denoted as BiGCN(SKP) and EBGCN(SKP), respectively.

For results of baselines, we implement BiGCN according to their public project\footnote{\url{https://github.com/TianBian95/BiGCN}} under the same environment. Other results of baselines are referenced from original papers \cite{DBLP:conf/aaai/KhooCQ020,DBLP:conf/emnlp/WeiXM19,DBLP:conf/acl/WongGM18}.

\section{Results and Analysis}
\subsection{Performance Comparison with Baselines}

\begin{table}[t]
  \centering
 \resizebox{0.98\columnwidth}{!}{
  \begin{tabular}{cccccc}
  \hline
   \multicolumn{6}{ c }{{\bf Twitter15}} 
        \\
    \hline
      \multirow{2}{*}{{Method}}   & \multirow{2}{*}{Acc.} & NR & F &  T & U  \\
    \cline{3-6}
   & & $F_1$ & $F_1$ & $F_1$ & $F_1$  \\
  \hline
  \multicolumn{1}{ l }{DTC 
   } & 45.5 & 73.3 & 35.5 & 31.7 & 41.5  \\
  \multicolumn{1}{l}{SVM-TS 
  }  & 54.4 & 79.6 & 47.2 & 40.4 & 48.3 \\
  \multicolumn{1}{l}{GRU-RNN 
  } &64.1 & 68.4 & 63.4 & 68.8 & 57.1   \\
  \multicolumn{1}{l}{SVM-TK
   } &
  66.7 & 61.9 & 66.9 & 77.2 & 64.5  \\
  \multicolumn{1}{ l }{RvNN 
  }  & 72.3 & 68.2 & 75.8 & 82.1 & 65.4 \\ 
  \multicolumn{1}{ l }{StA-PLAN 
  } & 85.2 & 84.0 & 84.6 & 88.4 & 83.7    \\ 
  \multicolumn{1}{ l }{BiGCN
  } & 87.1 & 86.0 & 86.7 &  91.3 & 83.6 \\
  \multicolumn{1}{ l }{\textbf{EBGCN} } & 
  \textbf{89.2} & \textbf{86.9} & \textbf{89.7} & \textbf{93.4} & \textbf{86.7} \\
  \hline
  \end{tabular}
  }
\resizebox{0.98\columnwidth}{!}{
  \begin{tabular}{cccccc}
    \hline
     \multicolumn{6}{ c }{{\bf Twitter16}} 
          \\
    \hline
    \multirow{2}{*}{{Method}}  & \multirow{2}{*}{Acc.} & NR & F &  T & U  \\
      \cline{3-6} 
     & & $F_1$ & $F_1$ & $F_1$ & $F_1$  \\
    \hline
    \multicolumn{1}{ l }{DTC 
    }  &46.5 & 64.3 & 39.3 & 41.9 & 40.3  \\
    \multicolumn{1}{l}{SVM-TS 
    }  & 54.4 & 79.6 & 47.2 & 40.4 & 48.3 \\
    \multicolumn{1}{l}{GRU-RNN 
    } & 63.6 & 61.7 & 71.5 & 57.7 & 52.7  \\
    \multicolumn{1}{l}{SVM-TK
     } &
    66.7 & 61.9 & 66.9 & 77.2 & 64.5  \\
    \multicolumn{1}{ l }{RvNN 
    }  & 72.3 & 68.2 & 75.8 & 82.1 & 65.4 \\ 
    \multicolumn{1}{ l }{StA-PLAN 
    } & 85.2 & 84.0 & 84.6 & 88.4 & 83.7    \\ 
    \multicolumn{1}{ l }{BiGCN
    } & {88.5} & {82.9} & {89.9} & {93.2} & {88.2}  \\
    \multicolumn{1}{ l }{\textbf{EBGCN}  } & 
    \textbf{91.5} & \textbf{87.9} & \textbf{90.6} & \textbf{94.7} & \textbf{91.0} \\
    \hline
    \\
\end{tabular}   
  }
\resizebox{0.98\columnwidth}{!}{
  \begin{tabular}{cccc}
        \hline
            \multicolumn{4}{ c }{\textbf{PHEME}} 
                \\
                \hline
        \multirow{1}{*}{Method}   & 
             { Acc. } & $\text{m}F_1$ & $\text{w}F_1$ \\
        \hline
        \multicolumn{1}{ l }{NileTMRG
          }
          & 36.0 & 29.7 & - \\ 
          \multicolumn{1}{ l }{BranchLSTM
              } & 31.4 & 25.9 & - \\
        \multicolumn{1}{ l }{RvNN 
        }
          & 34.1 & 26.4 & - \\ 
        
        \multicolumn{1}{ l }{Hierarchical GCN-RNN
        } &  35.6 & 31.7 & - \\
        \multicolumn{1}{ l }{BiGCN 
          }
        &  {49.2} & {46.7} & {63.2}  \\
        \multicolumn{1}{ l }{BiGCN(SKP) }
        &  {56.9} & {48.3} & {66.8}  \\
        \multicolumn{1}{ l }{\textbf{EBGCN} } &  \textbf{69.0} & \textbf{62.9} & \textbf{74.6}  \\  
        \multicolumn{1}{ l }{\textbf{EBGCN}(SKP) } & \textbf{71.5} & \textbf{57.5} & \textbf{79.1}  \\
        \hline
    \end{tabular} 
  }
\caption{Results (\%) of rumor detection.}
\label{tab:resultTwitter}
\end{table}

Table \ref{tab:resultTwitter} shows results of rumor detection on {\it Twitter15}, {\it Twitter16}, and {\it PHEME} datasets.
Our proposed model {\bf EBGCN} obtains the best performance among baselines.
Specifically, for {\it Twitter15}, {\bf EBGCN} outperforms state-of-the-art models 2.4\% accuracy and 3.7\% F1 score of unverified rumor. For {\it Twitter16}, our model obtains 3.4\% and 6.0\% improvements on accuracy and F1 score of non-rumor, respectively.
For {\it PHEME}, EBGCN significantly outperforms previous work by 40.2\% accuracy, 34.7\% $\text{m}F_1$ , and 18.0\% $\text{w}F_1$.

Deep learning-based ({\bf RvNN, StA-PLAN, BiGCN} and {\bf EBGCN}) outperform conventional methods using hand-crafted features ({\bf DTC}, \textbf{SVM-TS}), which reveals the superiority of learning high-level representations for detecting rumors.

Moreover, compared with sequence-based models {\bf GRU-RNN}, and {\bf StA-PLAN}, {\bf EBGCN} outperform them. It can attribute that they capture temporal features alone but ignore internal topology structures, which limit the learning of structural features. {\bf EBGCN} can aggregate neighbor features in the graph to learn rich structural features.

Furthermore, compared with state-of-the-art graph-based {\bf BiGCN}, {\bf EBGCN} also obtains better performance. We discuss the fact for two main reasons. First, {\bf BiGCN} treats relations among tweet nodes as reliable edges, which may introduce inaccurate or irrelevant features. Thereby their performance lacks robustness. \textbf{EBGCN} considers the inherent uncertainty in the propagation structure. In the model, the unreliable relations can be refined in a probability manner, which boosts the bias of express uncertainty. Accordingly, the robustness of detection is enhanced. Second, the edge-wise consistency training framework ensures the consistency between uncertain edges and the current nodes, which is also beneficial to learn more effective structural features for rumor detection.

Besides, {\bf EBGCN(SKP)} and {\bf BiGCN(SKP)}  outperforms {\bf EBGCN} and {\bf BiGCN} that use TF-IDF features in terms of Acc. and $\text{w}F_1$. 
It shows the superiority of word embedding to capture textual features.
Our model consistently obtains better performance in different text embedding. It reveals the stability of {\bf EBGCN}.

\begin{figure}[t]
  \centering
  \subfigure[The effect of edge inference]{\includegraphics[width=0.9\linewidth]{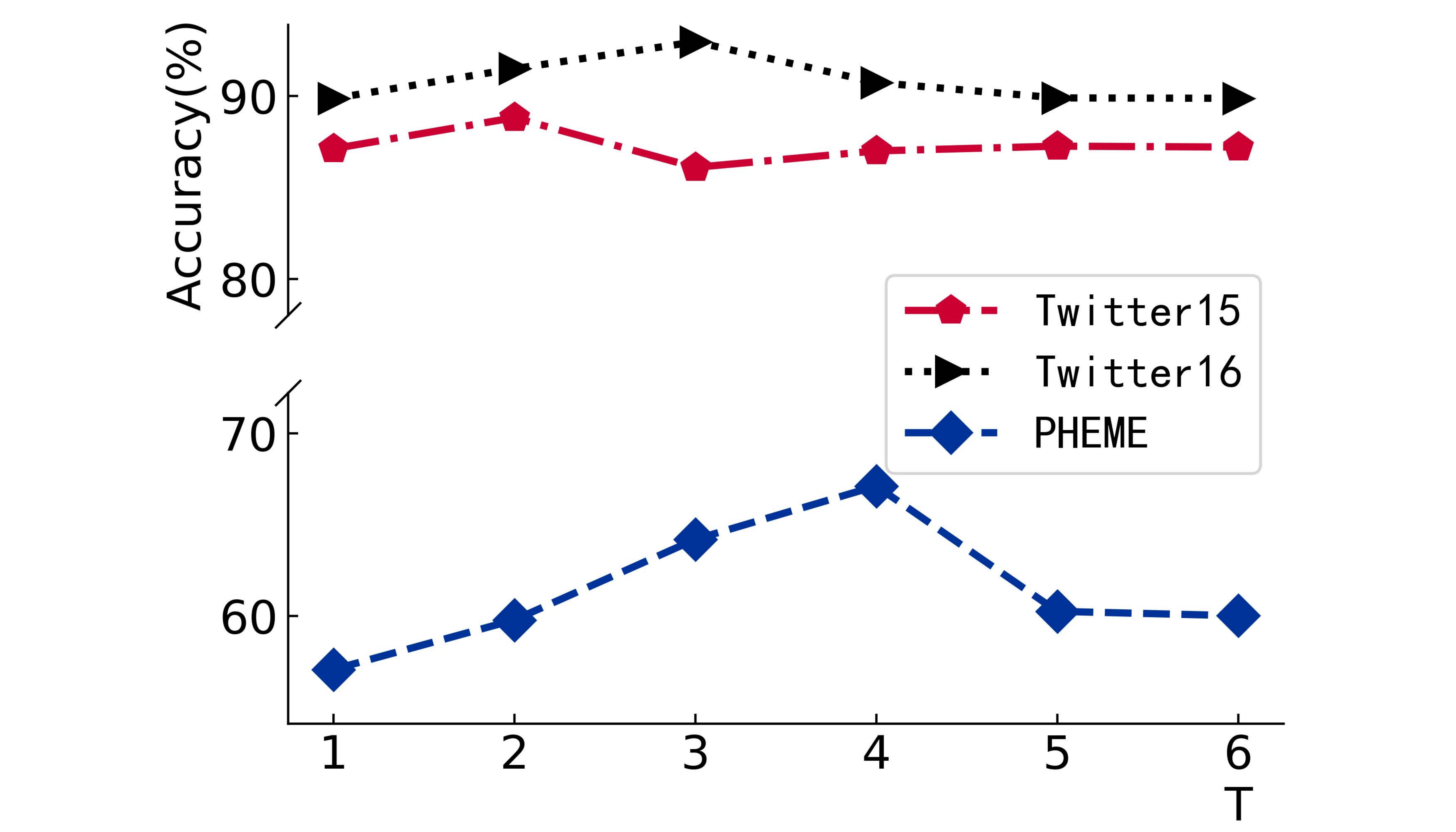}
  \label{fig:edge} }
  \subfigure[The effect of unsupervised relation learning loss]{\includegraphics[width=0.9\linewidth]{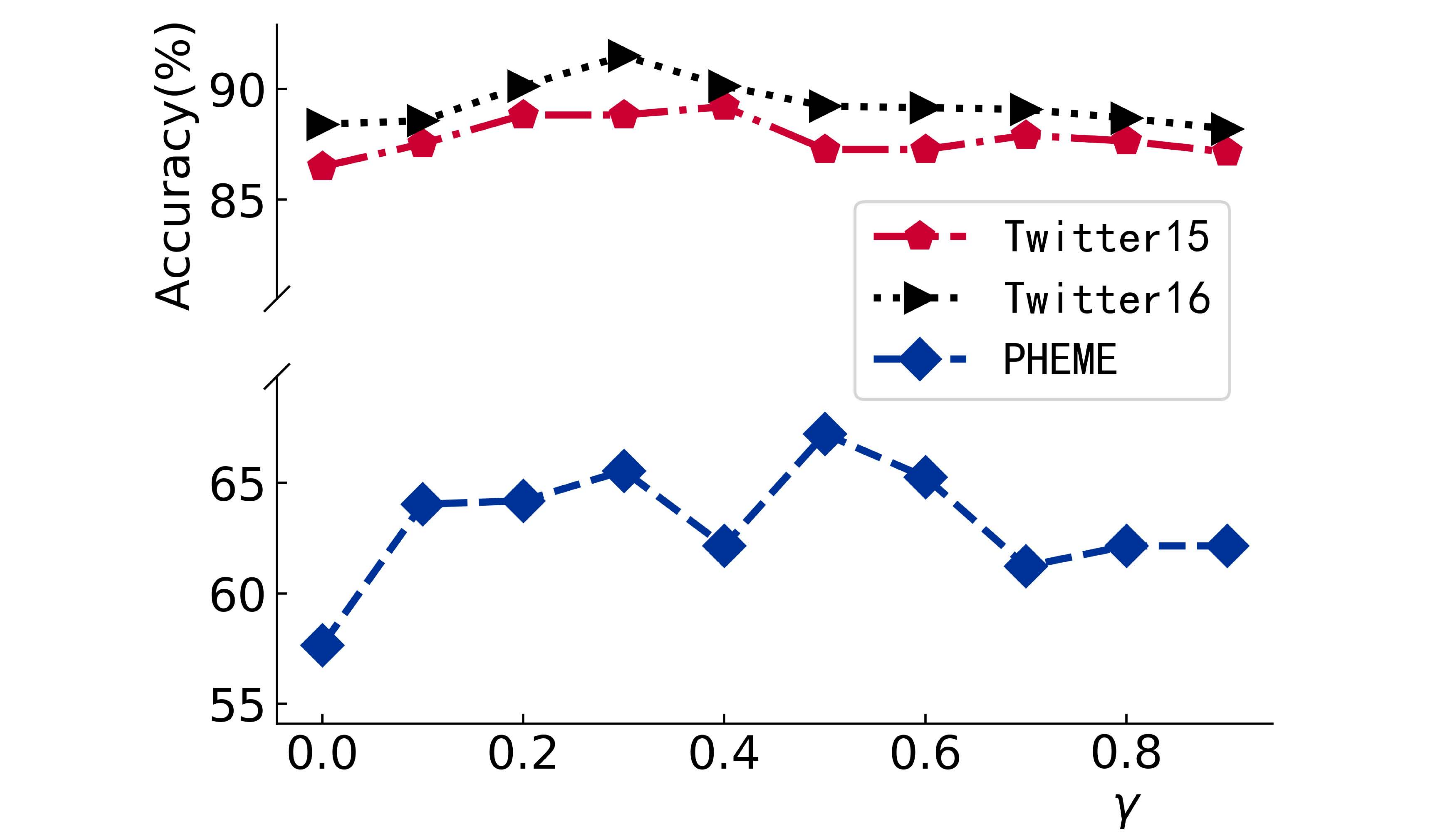}}
  \caption{Results of model analysis on three datasets. \label{fig:loss} } 
\end{figure}

\subsection{Model Analysis}
In this part, we further evaluate the effects of key components in the proposed model. 

\paragraph{The Effect of Edge Inference.}
The number of latent relation types $T$ is a critical parameter in the edge inference module.
Figure~\ref{fig:edge} shows the accuracy score against $T$. 
The best performance is obtained when $T$ is 2, 3, and 4 on {\it Twitter15}, {\it Twitter16}, and {\it PHEME}, respectively. 
Besides, these best settings are different. 
An idea explanation is that complex relations among tweets are various in different periods and gradually tend to be more sophisticated in the real world with the development of social media. 
The edge inference module can adaptively refine the reliability of these complex relations by the posterior distribution of latent relations. It enhances the bias of uncertain relations and promotes the robustness of rumor detection.

\paragraph{The Effect of Unsupervised Relation Learning Loss.}

The trade-off parameter $\gamma$ controls the effect of the proposed edge-wise consistency training framework. $\gamma=0.0$ means this framework is omitted. 
The right in Figure~\ref{fig:loss} shows the accuracy score against $\gamma$. When this framework is removed, the model gains the worst performance.
The optimal $\gamma$ is 0.4, 0.3, and 0.3 on {\it Twitter15}, {\it Twitter16}, and {\it PHEME}, respectively. 
The results proves the effectiveness of this framework.
Due to wily rumor producers and limited annotations of spread information, it is common and inevitable that datasets contains unreliable relations.
This framework can ensure the consistency between edges and the corresponding node pairs to avoid the negative features.

\begin{figure}[t]
  \centering
  \includegraphics[width=0.99\linewidth]{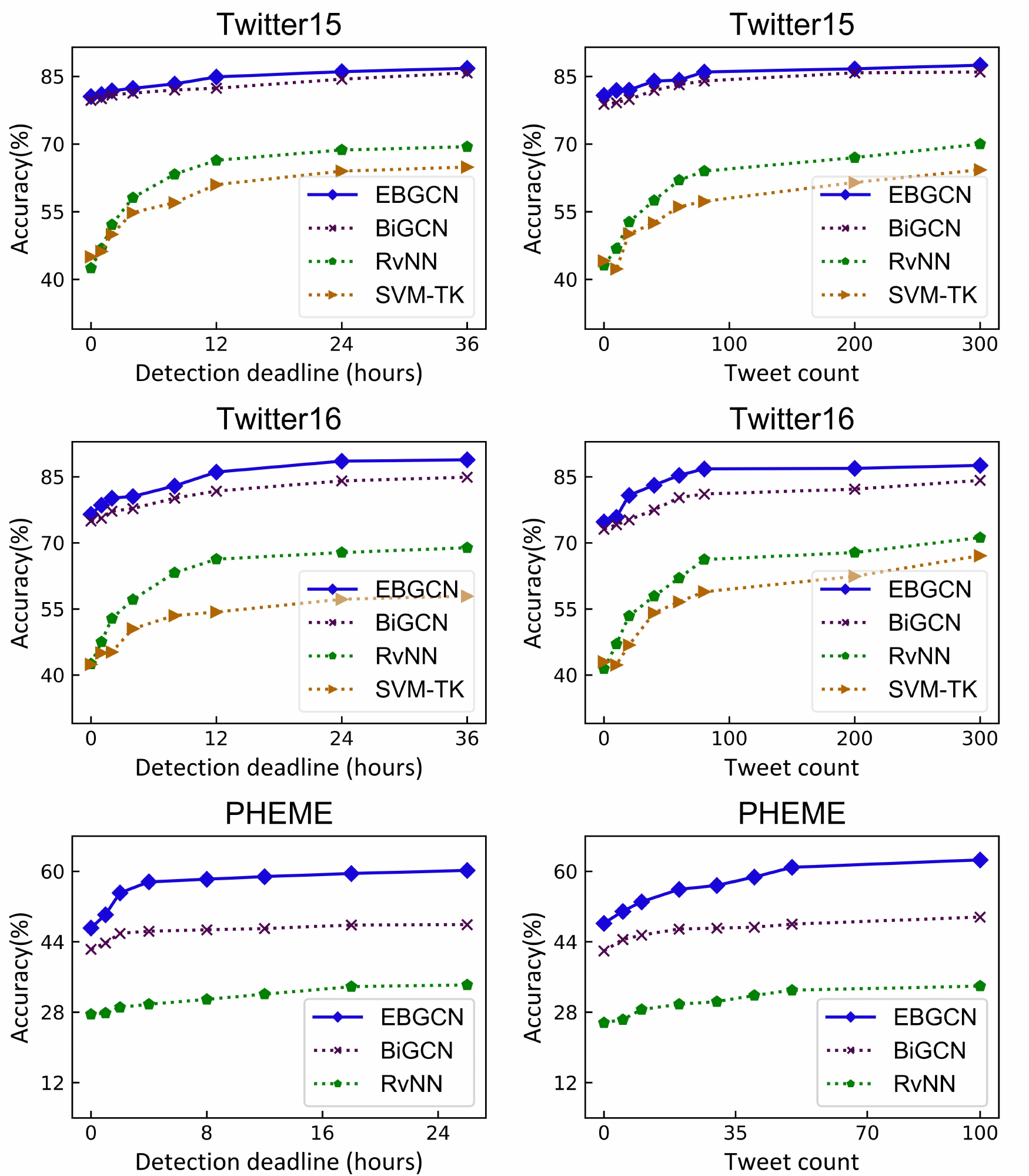}
    \caption{Performance of early rumor detection.}
\label{fig:early}
\end{figure}

\subsection{Early Rumor Detection}
Early rumor detection is to detect a rumor at its early stage before it wide-spreads on social media so that one can take appropriate actions earlier.
It is especially critical for a real-time rumor detection system.
To evaluate the performance on rumor early detection, we follow \cite{DBLP:conf/acl/WongGM18} and control the detection deadline or tweet count since the source tweet was posted. The earlier the detection deadline or the less the tweet count, the less propagation information can be available.

Figure~\ref{fig:early} shows the performance of early rumor detection.
First, all models climb as the detection deadline elapses or tweet count increases. 
Particularly, at each deadline or tweet count, our model {\bf EBGCN} reaches a relatively high accuracy score than other comparable models.

Second, compared with {\bf RvNN} that captures temporal features alone and {\bf STM-TK} based on handcrafted features, the superior performance of {\bf EBGCN} and {\bf BiGCN} that explored rich structural features reveals that structural features are more beneficial to the early detection of rumors.

Third, {\bf EBGCN} obtains better early detection results than {\bf BiGCN}. It demonstrates that {\bf EBGCN} can learn more conducive structural features to identify rumors by modeling uncertainty and enhance the robustness for early rumor detection.

Overall, our model not only performs better long-term rumor detection but also boosts the performance of detecting rumors at an early stage.

\begin{figure*}[t]
  \centering
  \includegraphics[width=0.8\linewidth]{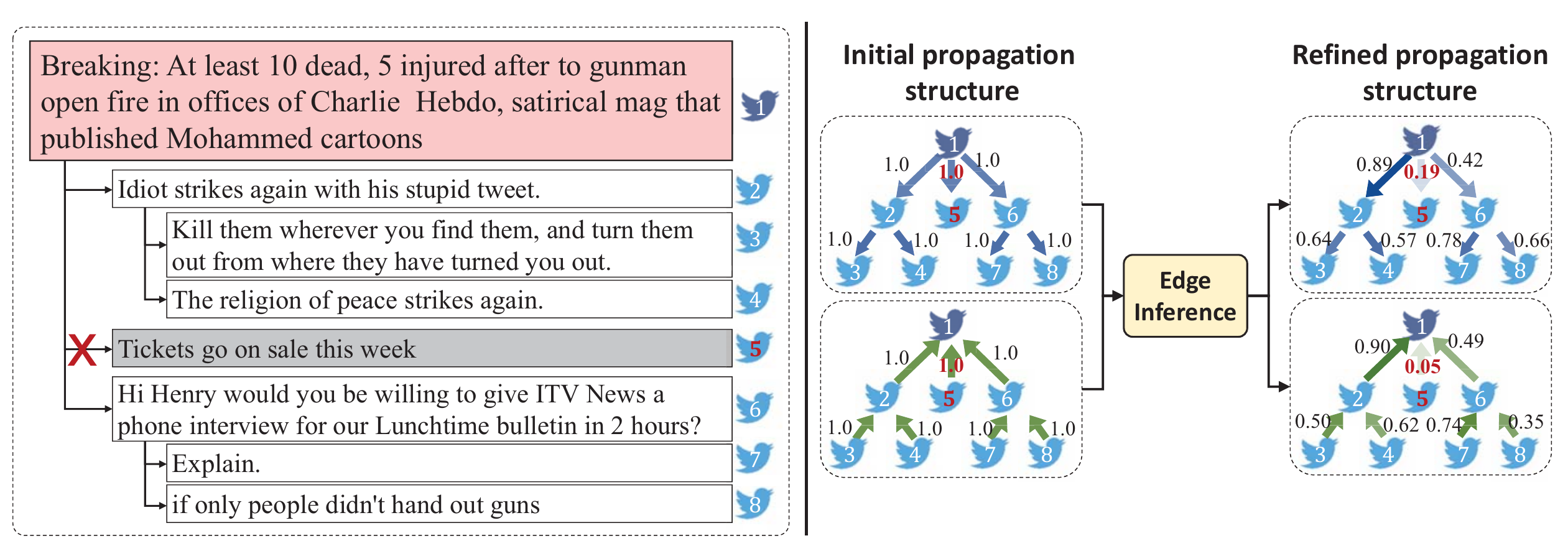}
  \caption{The case study. Left shows a false rumor sampled from {\it PHEME}. The gray-highlighted tweet is the irrelevant one towards this rumor propagation but included in. Right is the constructed directed graphs in top-down and bottom-up directions based on the propagation structure. Our model iteratively adjusts the weights of edges in each graph to strength the effect of reliable edges and weaken the effect of unreliable edges. }
  \label{fig:case}
\end{figure*}

\subsection{The Case Study}

In this part, we perform the case study to show the existence of uncertainty in the propagation structure and explain why EBGCN performs well. We randomly sample a false rumor from {\it PHEME}, as depicted in Figure~\ref{fig:case}.  The tweets are formulated as nodes and relations are modeled as edges in the graph, where \textit{node} 1 refers to the source tweet and \textit{node} $2\text{-}8$ refer to the following retweets.  

As shown in the left of Figure~\ref{fig:case}, we observe that \textit{tweet} 5 is irrelevant with \textit{tweet} 1 although replying, which reveals the ubiquity of unreliable relations among tweets in the propagation structure and it is reasonable to consider the uncertainty caused by these unreliable relations.  

Right of Figure~\ref{fig:case} indicates constructed graphs where the color shade indicates the value of edge weights. The darker the color, the greater the edge weight. The existing graph-based models always generate the representation of \textit{node} 1 by aggregating the information of its all neighbors (\textit{node} 2, 5, and 6) according to seemingly reliable edges.  However, \textit{edge between node 1 and 5} would bring noise features and limit the learning of useful features for rumor detection. Our model EBGCN successfully weakens the negative effect of this edge by both the edge inference layer under the ingenious edge-wise consistency training framework. Accordingly, the model is capable of learning more conducive characteristics and enhances the robustness of results.

\section{Conclusion}
In this paper, we have studied the uncertainty in the propagation structure from a probability perspective for rumor detection. Specifically, we propose Edge-enhanced Bayesian Graph Convolutional Networks (EBGCN) to handle uncertainty  with a Bayesian method by adaptively adjusting weights of unreliable relations. Besides, we design an edge-wise consistency training framework incorporating unsupervised relation learning to enforce the consistency on latent relations. Extensive experiments on three commonly benchmark datasets have proved the effectiveness of modeling uncertainty in the propagation structure. EBGCN significantly outperforms baselines on both rumor detection and early rumor detection tasks. 


\bibliographystyle{acl_natbib}
\bibliography{anthology,acl2021,rumor}


\end{document}